\pdfoutput=1

\documentclass[11pt]{article}

\usepackage[]{acl}
\usepackage{xcolor}
\usepackage{times}
\usepackage{latexsym}

\usepackage[T1]{fontenc}
\usepackage{multirow}
\usepackage[utf8]{inputenc}
\usepackage{url}
\usepackage{microtype}
\usepackage{times}
\usepackage{helvet}
\usepackage{courier}
\usepackage{booktabs}
\usepackage{amssymb}
\usepackage[noend]{algpseudocode}   
\usepackage{hyperref}
\usepackage{algorithmicx,algorithm}
\title{Challenges for Open-domain Targeted Sentiment Analysis}

\usepackage{epsfig}

\author{
    Yun Luo \textsuperscript{\rm1,2},
    Hongjie Cai \textsuperscript{\rm3},
    Linyi Yang \textsuperscript{\rm1,2},
    Yanxia Qin \textsuperscript{\rm4}, 
    Rui Xia \textsuperscript{\rm3}, 
    Yue Zhang \textsuperscript{\rm1,2}
    \\
    \textsuperscript{1} School of Engineering, Westlake University, Hangzhou, China. \\
    \textsuperscript{2} Institute of Advanced Technology, Westlake Institute for Advanced Study, Hangzhou, China.  \\
    \textsuperscript{3} School of Computer Science and Engineering, Nanjing University of Science and Technology, Nanjing,
China. \\
    \textsuperscript{4} 	School of Computing, National University of Singapore, Singapore. \\
    \texttt{\{luoyun, yanglinyi, zhangyue\}@westlake.edu.cn}\\
    \texttt{\{hjcai, rxia\}@njust.edu.cn} \\
    \texttt{yxqin@nus.edu.sg
}\\
}
\begin{document}
\maketitle
\begin{abstract}
Since previous studies on open-domain targeted sentiment analysis are limited in dataset domain variety and sentence level,  we propose a novel dataset consisting of 6,013 human-labeled data to extend the data domains in topics of interest and document level. Furthermore, we offer a nested target annotation schema to  extract the complete sentiment information in documents, boosting the practicality and effectiveness of open-domain targeted sentiment analysis. Moreover, we leverage the pre-trained model BART in a sequence-to-sequence  generation method for the task. Benchmark results show that there exists large room for improvement of open-domain targeted sentiment analysis. Meanwhile, experiments have shown that challenges remain in the effective use of open-domain data, long documents, the complexity of target structure, and domain variances.
\end{abstract}

\section{Introduction}
\noindent

\begin{figure}[t]
      \centering
      \includegraphics[width = \hsize]{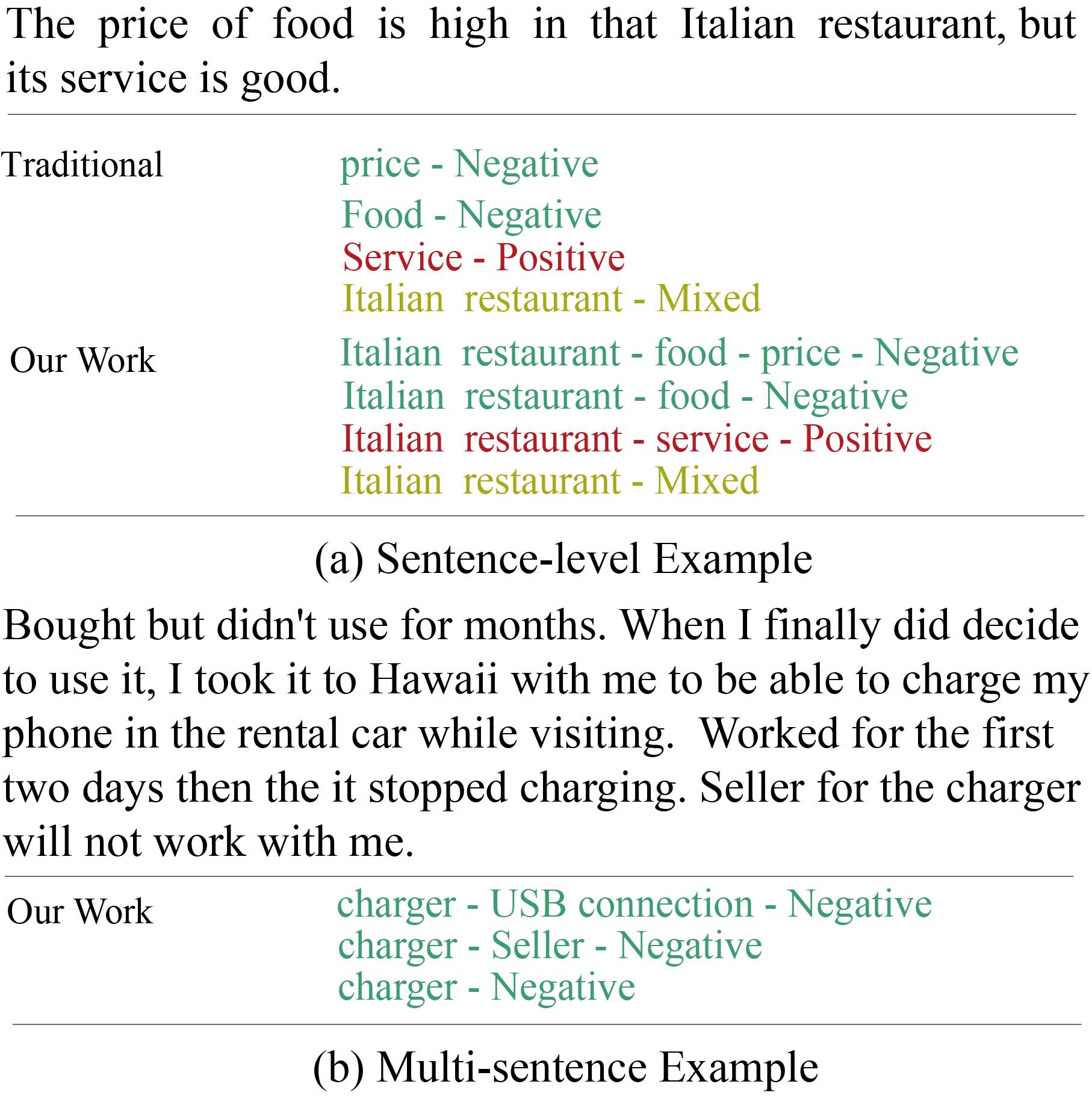}
      \caption{Traditional open-domain targeted sentiment analysis (Traditional in the figure) and our work.}
      \label{ourwork}
   \end{figure}

Open-domain targeted sentiment analysis refers to the task of extracting entities and sentiment polarities (e.g. positive, negative, neutral) towards them in free texts \cite{Mitchell2013} (Figure \ref{ourwork}). It has received much research attention due to wide applications to market prediction, recommendation system, product selection, public opinion surveillance. For example, a business might be interested in monitoring the mentioning of itself or its products and services from all media sources, and an investment fund can be interested in learning the sentiment towards a range of open-ended topics that can potentially be influential to market volatilities. Ideally, the task requires algorithms to process open-domain texts from different genres such as news, reports and tweets. For each domain,  topics and opinion expressions can be highly different.

As shown in Figure \ref{ourwork} (a), existing research on open-domain targeted sentiment has focused on a sentence-level setting  \cite{Mitchell2013}, where different models have been proposed to extract or tag text spans as the mentioned targets, assigning sentiment polarity labels (i.e., positive, negative and neutral) on each extracted span. Both pipeline methods \cite{Mitchell2013,Zhang2015,Hu2020}  and joint methods \cite{Mitchell2013,Zhang2015,Ma2018,Li2018b,Zhou2019,Song2019,Pingili2020,Hu2020}  have been considered, with the former taking separate models for opinion target extraction and target sentiment classification, and the latter using a single model for solving both subtasks. The current state-of-the-art results \cite{luo2020grace} has been achieved by using  pre-trained model BERT \cite{devlin2018bert}.


   \begin{table*}[thpb] \small
\centering
\begin{tabular}{lcccccccccccccccc}   
\hline
\hline  
\textbf{Domain} &\textbf{\#Doc }& \textbf{\#T}& \textbf{\#P} & \textbf{\#N}&\textbf{\#M} &\textbf{F1} &\textbf{Kappa} & \textbf{\#S} & \textbf{\#Tok} & \textbf{\#AT}\\
\hline   
\textbf{Books} &986&2,470&1,624&542&304&59.06&62.94&7.59&109.10&2.50\\
\textbf{Clothing} &928&1,555&1034&299&222&60.51&70.25&4.54&44.12&1.67\\
\textbf{Restaurant} &940&4,739&3,457&828&454&57.44&63.88&10.08&116.63&5.03\\
\textbf{Hotel} &1,029&3,436&3,165&154&117&72.07&75.82 &5.24&55.63&3.33\\
\textbf{News}& 936 &2,725 &1,358&1,254&113&75.34&76.08&12.53  & 175.72&2.91\\
\textbf{PhraseBank} &1,194  &1,481 &1,006&464&11&75.04&77.49&1.00&23.30&1.23 \\
\hline
\hline
\end{tabular}

\caption{Details for our proposed datasets, include the number of documents (\#Doc) and targets (\#T) in each domain, the number of Positive (\#P), Negative (\#N), Mixed (\#M) sentiment labels, micro-F1 scores of annotator agreement (F1 for micro F1 score, henceforth), the average number of sentence (\#S), tokens (\#Tok), and targets (\#AT)  each domain, respectively.}
\label{Table1}
\end{table*}
   
Existing work, however, is limited in several aspects. First, it is constrained by the use of relatively small datasets from \citet{Mitchell2013} and \citet{pontiki2014,pontiki2015,pontiki2016}, which are confined to the restaurant review, laptop review and twitter domains. As a consequence, the state-of-the-art methods that perform well in these closed domains come to be manifest in performance problems when faced with open-domain data \cite{Orbach2020}. Recent availability of representation models pre-trained on diverse text domains \cite{devlin2018bert,radford2019language,lewis-etal-2020-bart} allows us to investigate open-domain targeted sentiment in more practical and realistic settings.

Second, existing work considers open-domain targeted sentiment analysis only at the sentence level. However, text sources in the open domain are typically in the form of documents, such as a piece of news, or a product review. Note that documents refer to the complete text from a text author such a complete review, or a complete news, which  means the length of text varies,  meeting the characteristic of open-domain. Sentence-level sentiment models fail to give accurate information due to lack of co-reference and discourse knowledge. Take the simple sentence ``{ \textit{It is quite useful in helping me with the housework.}}'' from the dataset of \citet{Mitchell2013} for example, the gold-standard target entity is represented by the span ``{\it it}''. However, significant post-processing can be necessary to correctly identify the true sentiment polarity on the target entity, which involves co-reference  resolution and mention-level polarity information integration.

Third, complex relations are not fully extracted in the previous work, which just extracts opinion targets separately. 
For example, `The \textit{price} of the \textit{food} is high  in that \textit{Italian restaurant}', the relationships of \textit{price}, \textit{food} and \textit{Italian restaurant} are not implied in the previous datasets. Although some work extracts the target, aspect and sentiment at the same time \cite{Yang2019,Saeidi2016}, it is still limited in the extensibility, having restricted the schema of target expression e.g. \textit{food-price-Negative} which is three-tuple failing to indicate \textit{Italian restaurant}.


 To address the above issues, we consider open-domain targeted sentiment analysis at the document level with a variety of domains. A contrast between our dataset and traditional open-domain targeted sentiment analysis is shown in Figure \ref{ourwork}. In particular, for increasing diversity, our data are sourced from six different domains with three different linguistic genres. To address the limitation on span-based target representation, we define the problem of open-domain targeted sentiment as a fully end-to-end task, where the input is a document and the output is a list of mentioned target entities with their sentiment polarities.


While pre-trained sequence-to-sequence  models are useful for solving our task,  results  show that there is a large gap for further improvement. Challenges exist in  the effective use of open-domain data, long documents, the complexity of target structure, and domain variances.   To our knowledge, we are the first to consider the open-domain targeted sentiment analysis in the  document-level  setting. The resulted dataset has been released at Google Drive. \footnote{\href{https://drive.google.com/drive/folders/10cZW636Rp8iQZBTxER72rvjo7JiC-QDR?usp=sharing}{https://drive.google.com/drive/folders/10cZW636Rp8iQZ BTxER72rvjo7JiC-QDR?usp=sharing}}

      \begin{figure*}[t]
      \centering
      \includegraphics[width = 0.9\hsize]{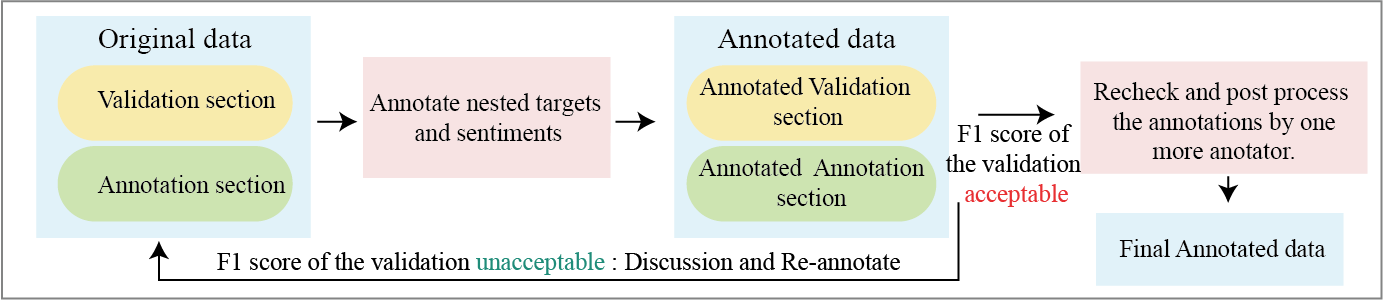}
      \caption{Annotation procedure of our proposed dataset.}
      \label{procedure}
   \end{figure*}

\section{Dataset}
  Our proposed dataset contains six domains, including book reviews, clothing reviews, restaurant reviews, hotel reviews, financial news and social media data (PhraseBank). The details of data sources are shown in Appendix B.

\subsection{Annotation Schema}
 Considering that targets can have fine-grained levels of specificity (e.g., restaurant-food-price), we denote sentiment targets with tuples, where all the targets and their relations are extracted in a nested data structure \ref{ourwork}. To allow better document-level representation and avoid noise, we adopt the \{Positive, Negative, Mixed\} sentiment schema \cite{Orbach2020}.

\subsection{Annotation Procedure}
The procedure of annotation is shown in Figure \ref{procedure}. Each domain is distributed three different annotators, who are trained before making annotation.  The data is divided into the annotation and validation sections -- the former is allocated to one of the annotators, and the latter is annotated by at least two annotators. After annotation, we calculate the average  micro F1 score of each two annotators to check annotation agreements on the validation section. The F1 score is calculated in phrase level for the reason we consider the relations of target components in the evaluation procedure, similar to \citet{kim-klinger-2018-feels}. If the F1 score does not reach an acceptable level, we discuss about the issues and revise the annotation guidelines when necessary, 
after that the data are re-annotated. If the F1 score reaches an acceptable level, the data are re-checked by one more individual.  The details of the final annotation rules are shown in Appendix A. We also calculate the Kappa $\kappa$ value of the labeled data in word level to evaluate the agreement (in Table 1), all of which are higher than 60\%.

Considering the complexity of the nested target structure, we use a loose-match score replacing the exact-match score in the calculation of the F1 score, which is also used in our experimental evaluation. The exact-match score means that each labeled target is assigned correct score 1.0 only if all the components and the sentiment are the same with the golden text. But in loose-match score for each target if the sentiment is correct, we calculate the ratio of overlapped nests in labels and the golden text, and if the ratio reaches acceptable levels, we assign it with corresponding scores. The loose-match score is chosen for the annotation because the components of nested targets tend to have similar sentiments.  For example in Figure \ref{ourwork} (a), in \textit{Italian  restaurant - food - price - Negative},  the target components \textit{food} and \textit{Italian restaurant} also tend to have negative polarities for \textit{high price}. The acceptable levels we set 0.5 and 0.66 with the corresponding score 0.5, 1.0.

\subsection{Analysis and Statistics}
 Table \ref{Table1} shows the statistics in each domain of our dataset. First, the numbers of documents are roughly the same for each domain, with all domains having more than 900 documents. Second, the average number of sentences is the smallest in the \textbf{PhraseBank} domain which is one feature of the \textbf{PhraseBank}, and the largest in the \textbf{News} domain. The average number of targets is the largest in \textbf{Restaurant} reviews implying the difficulty in this domain is the largest. Third, label imbalance exists in the dataset, with positive sentiments being the dominant. We did not deliberately control the label distribution, to keep it as close to practical situations as feasible (similar to \citet{pontiki2014,pontiki2015,pontiki2016}).




     \begin{figure*}[t]
      \centering
      \includegraphics[width = 0.6 \hsize]{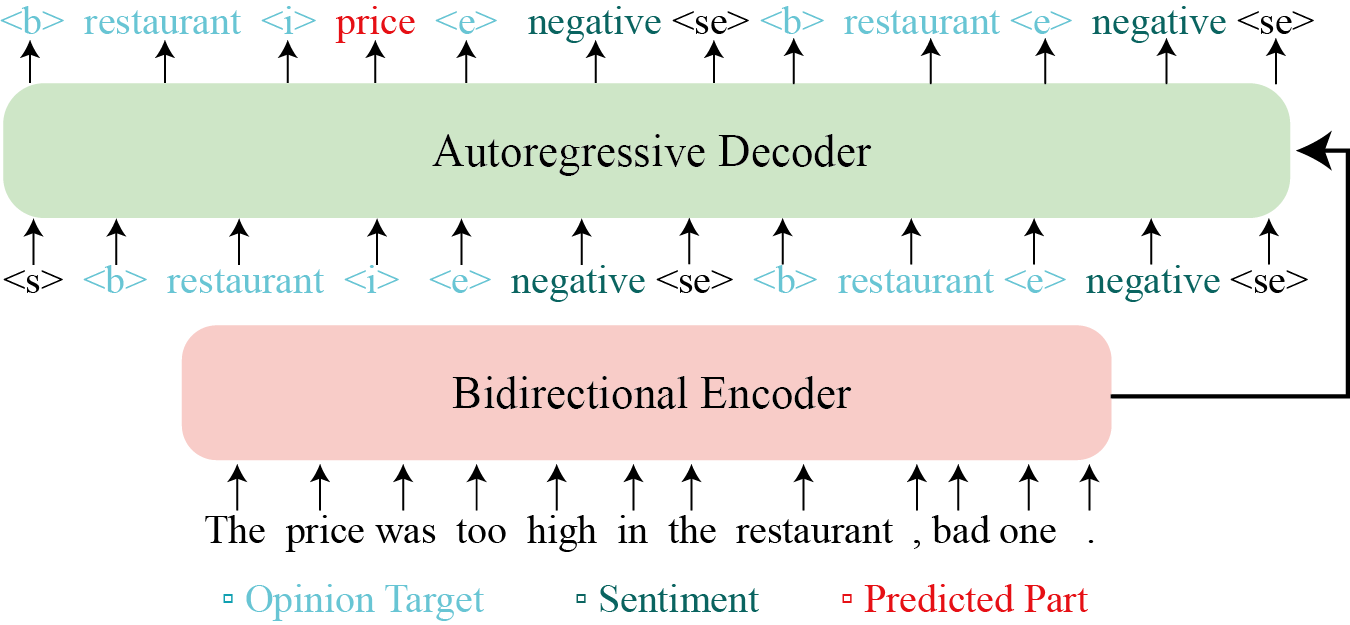}
      \caption{Pre-trained model for generation method in open-domain targeted sentiment analysis task.}
   \end{figure*}
\section{Approach}
In our schema, the nested opinion targets are in a structure that involves the relations of each component and inference of implicit targets, which can be challenging for traditional structured prediction models \cite{Mitchell2013,Zhang2015}.   Neural sequence-to-sequence modeling provides a useful solution \cite{vinyals2015order}, and we take BART \cite{lewis-etal-2020-bart}  as the sequence-to-sequence framework, which is a denoising autoencoder for pre-training sequence-to-sequence models based on Transformer \cite{vaswani2017attention}. BART  has shown to be particularly effective in tasks of  text summarization, machine translation, information retrieval and sequence generation \cite{lewis-etal-2020-bart,liubart,9390683,liu2021solving,yan-etal-2021-unified}.


\subsection{Model}
We consider both the joint task of open-domain targeted sentiment analysis and its subtasks. Formally our model takes $X = [x_1,x_2,...,x_n]$ as inputs, and output a target sequence  $Y_t$. For target sentiment classification, the output is $Y_s$, a token sequence generated using a given template. Both of $Y_s$ and $Y_t$ are formulated as $[y_0,y_1,...,y_n]$ where $y_0$ is the beginning token of BART.

\subsubsection{Opinion Target Extraction}
 For {opinion target extraction}, the target sequence
 $[y_1,...,y_m]$ (not includes the beginning token for BART) is a target list $[t_1, t_2,..,t_l]$. Each element is $t_j = [e_b,...,e_i,...,e_e]$ where  $e_b$,  $e_e$ are the beginning and ending token of each target respectively, and $e_i$ is the token to separate the nest structure of targets. For instance, given the input `\textit{The food in this restaurant is awful}', the output is    $[e_b,\ restaurant,\ e_i, \ food,  \ e_e, $ $ \ e_b,\ restaurant, \ e_e]$.

\subsubsection{Target Sentiment Classification}
For {target sentiment classification}, we set a target set for each document $T = \{t_1, t_2,..,t_{|L|}\}$ where $|L|$ is the number of targets for each document in the dataset and the sentiment polarity set $P = \{p_1,p_2,...,p_{|C|}\}$ where $|C|$ is the number of sentiment polarity in the task. Each element $t_j = e_b,...,e_i,...,e_e$ is in the same format mentioned above. Similar to \citet{liu2021solving}, we create the templates $\mathbf{T}_{t_j,p_k} ={w_1,w_2,...,w_l} = [t_j + p_k,e_{se}]  $ (e.g. $[e_b,\ restaurant,\ e_i, \ food,\ e_e, \ positive, \ e_{se}]$). For a given target set, we can obtain a list of templates $\mathbf{T}_{t_j} = [T_{t_j,p_1},T_{t_j,p_2},...,T_{t_j,p_{|C|}}]$, and feed the template sets into fine-tuned pre-trained generative language model to assign a score to each template $\mathbf{T}_{t_j,p_k} = {w_1,w_2,...,w_l}$:
\begin{equation}
    f (\mathbf{T}_{t_j,p_k}) = \sum _{i = 1} ^{l} log P({w_i|w_{1,i-1},X})
\end{equation}
    
We choose the sentiment polarity  with the largest score for the target $t_j$.

\subsubsection{Open-domain Targeted Sentiment Analysis}
 For {{open-domain targeted sentiment analysis}}, the target sequence $[y_1,...,y_m]$ (not includes the beginning token for BART) is a target list $[t_1, t_2,..,t_l]$. Each element is $t_j = [e_b,...,e_i,...,e_e,s_j,e_{se}]$, where  $e_b$,  $e_e$, $e_{se}$ are the beginning, ending tokens of each target, and ending token of sentiment respectively. $e_i$ is to separate the nest structure of targets, $s_j$ is the sentiment towards this target. For example, given the input `\textit{The food  is too awful}', the model output is    $[e_b, \ food, \ e_e, \ negative, \ e_{se}]$.

\begin{table*}[tph]\small

\centering
\begin{tabular}{c|ccc|c|ccc|ccc}
\hline
\hline
\multirow{2}{*}{\textbf{Domain}} & \multicolumn{3}{c|}{\textbf{OTE}}         & \multicolumn{1}{c|}{\textbf{TSC}}       & \multicolumn{3}{c|}{\textbf{OTSA}}  &\multicolumn{3}{c}{\textbf{OTSA-Single}}     \\
& \textbf{Precision} & \textbf{Recall} & \textbf{F1} & \textbf{Precision} & \textbf{Precision} & \textbf{Recall} & \textbf{F1}& \textbf{Precision} & \textbf{Recall} & \textbf{F1} \\
\hline
\textbf{Books}                   &   56.84    &  38.12      &    45.63           &  73.85   &    40.65   &    26.25   &  31.90 & {43.02}         &   29.17    &     {34.76}       \\
\textbf{Clothing}                &     62.93      &  47.20    &  53.94         &    83.55     &  49.36       & 38.32       &  43.14   &     {60.67}     &    {41.66}   &        {49.40}        \\
\textbf{Restaurant}              &     47.11     & 25.46       & 33.05           &  83.26  &       32.00   &    15.44    &  20.82   &        {35.98}   &    12.99    &    19.08         \\
\textbf{Hotel}                    &  68.85       &  44.14   &     53.79          &    95.69     &     50.39    &   29.38    & 37.12     &       47.67    &   26.64     & 34.17         \\
\textbf{News}                    &  23.16         &  10.93      &    14.85     &   69.94    &       20.23    &11.33           & 14.52          &     {18.57}      &    9.90    &    12.91             \\
\textbf{PhraseBank}              &   63.28     & 54.10     &     58.32        &  91.48 &       60.67  &    52.62     & 56.35   &{58.92}&{52.05}&{55.27}           \\
\hline
\textbf{Avg}                           &     53.70      &    36.66       &   43.26     & 82.96 &    42.21       &  28.89      &    33.98 &  44.13        &    28.73   & 34.27 \\
\hline
\hline
\end{tabular}
\caption{Experimental results  (OTE for opinion target extraction task, TSC for target sentiment classification task and OTSA for results of open-domain targeted sentiment analysis on the multi-domain setting; OTSA-Single for results of open-domain targeted sentiment analysis on the single-domain setting).}
\label{result4}
\end{table*}
 


\subsubsection{Training}
In opinion target extraction and open-domain targeted sentiment analysis, the gold outputs are given directly as a token list $Y_t$. For target sentiment classification, gold texts $Y_s$ are generated for each target with a gold polarity by using templates.

Given a sequence input $X$, we feed the input $X$ into BART encoder to obtain the hidden states: 
\begin{equation}
    \mathbf{h}^{encoder} = BARTEncoder(X)
\end{equation}

At  the $i-th$ step of the BART decoder, the generated output tokens $y_{1:i-1}$ are taken as inputs to yield a representation  
\begin{equation}
    \mathbf{h}^{decoder}_i = BARTDecoder(h^{encoder},y_{1:i-1})
\end{equation}

 The loss  function for the training instance $(X,Y_t)$ or $(X,Y_s)$ is formulated as 
\begin{equation}
    \mathcal{L} = - \sum _{i = 1} ^{m} log P({y_i|y_{1,i-1},X})
\end{equation}

\section{Experiments}
We conduct experiments for verifying the influence of the open-domain data, the document length, the complex target structure and the model structure in open-domain targeted sentiment analysis. 

\subsection{Experimental Settings}
We perform experiments using the official pre-trained BART model provided by Huggingface\footnote{https://huggingface.co/facebook/bart-base}. The maximum input sequence length is 512, and the maximum  output sequence length is 100. We split our dataset into training/validation/testing sets in the same ratio of 7:1:2 for all tasks.  The best model configuration is selected according to the highest performance on the validation set. In particular, the batch size 4, learning rate is initialized as 1e-4, our model is trained for 20 epochs. The experiments include:

\textbf{Multi-domain and single-domain settings.}
We first mix up the data on the six domains and fine-tune the BART model over a multi-domain setting before testing the trained model on the mixed data and the data in each domain, respectively.  Then we carry out experiments over the single-domain setting, by training the model on a single domain and test the model on the corresponding test data.

\textbf{Test on complex nested target structure.}
For exploring the influence of complex nested target structure, we try to mix the datasets and split the data  w.r.t. the number of target nests.   The statistics of the number of targets  with different numbers of nests in each domain is shown in Table \ref{Table1} (last 4 columns).  We train and test the model on each data split of different numbers of nests (1-nest, 2-nest and 3-nest) respectively.

\textbf{Out-of-domain test.} Models for open-domain targeted sentiment analysis are expected to learn sufficient knowledge about various domains and be applied to unseen domains for open-domain requirements. We design 5-1 (1-1) out-of-domain tests, using training data on five (one) domains to train the model, and testing the model on another domain. 

\textbf{Pipeline model.} In order to evaluate the performance of the pipeline model, we train the model of opinion target extraction and target sentiment classification on  each domain separately and test on the model pipeline.

\subsection{Overall Results}
First, the loose-match evaluation scores  of the multi-domain setting experiment on test mixed data are precision 41.40, recall 25.10, F1 31.25, relatively higher than the exact match evaluation score (precision 19.13, recall 17.66, F1 21.98).  The values of loose-match evaluation scores provide evidence that there exist much room for improvement in open-domain targeted sentiment analysis, comparing with the F1 score reported by the previous traditional work \cite{Hu2020} where the F1 scores of \textbf{LAPTOP},  \textbf{REST}, \textbf{Twitter} are  68.06, 57.69 and 74.92, respectively.  Meanwhile, the F1 score of Transformer model on mixed test data is only 3.76, which indicates the significance of using pre-trained models for the task.

\begin{figure}[t]
      \centering
      \includegraphics[width=1.0\hsize]{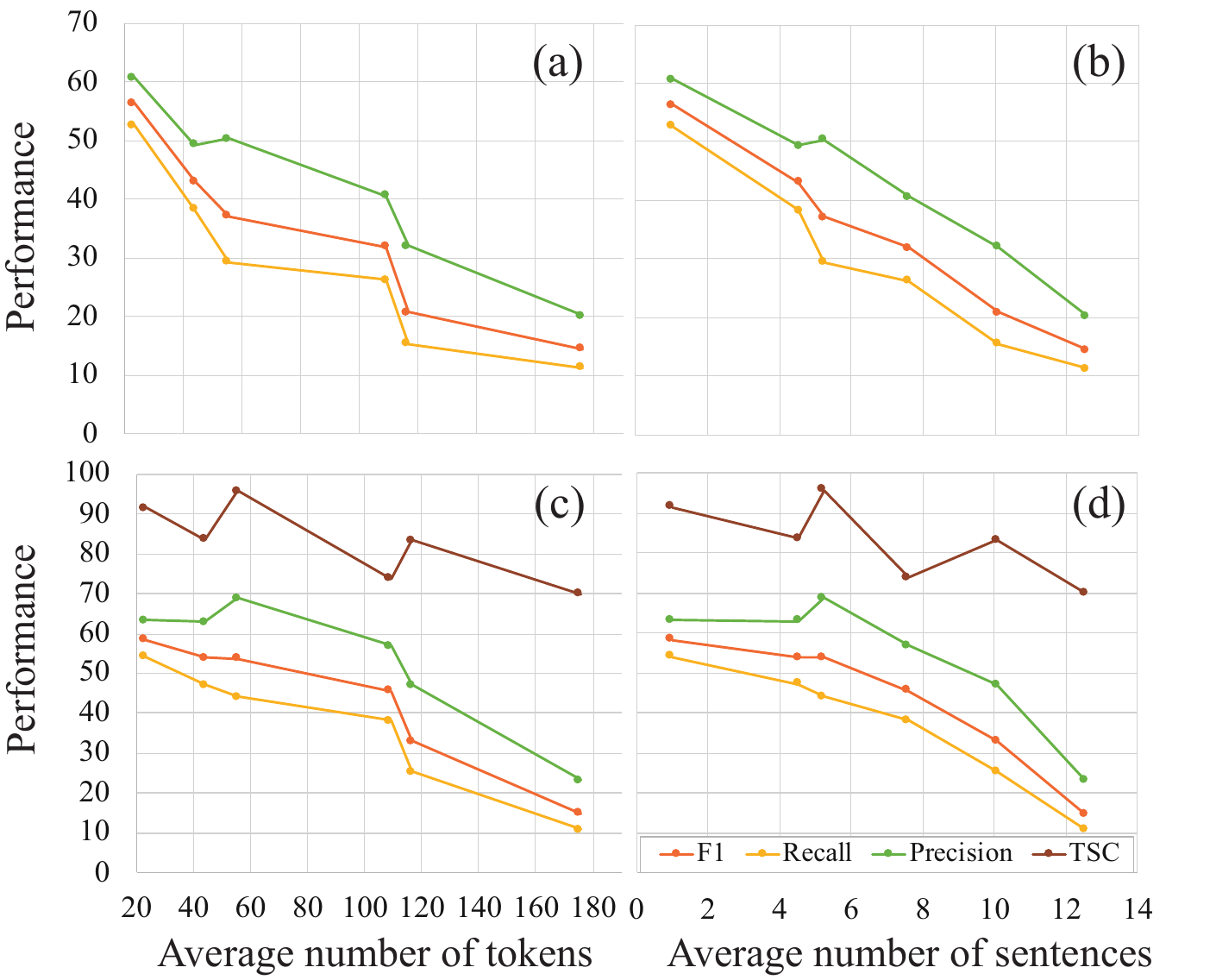}
      \caption{The relation between the document length and the performance on the multi-domain setting. (a) (b) for open-domain targeted sentiment analysis; (c) (d) for opinion target extraction and target sentiment classification.}
      \label{tokens}
  \end{figure}
  
Second, the results of the multi-domain setting trained BART model of each domain are shown in Table \ref{result4} (first seven columns).  The performance of open-domain targeted sentiment analysis on \textbf{Books} (31,90), \textbf{Restaurant} (20.82) and \textbf{News} (14.52) domains are relatively the weakest. This could be due to different factors including the size of documents, domains, and target structure, which are analyzed in Section 4.3, 4.4, and 4.5, respectively. 

Third, it is worth noting that the average recall values (36.66 and 28.89) for opinion target extraction and  open-domain targeted sentiment analysis are all lower than the precision (53.70 and 42.21). It suggests that the model tends to output more correct targets and sentiments, but fails to identify all the targets and sentiments.  Then, by comparing the results on opinion target extraction and target sentiment classification, the precision of the latter task (82.96) is strongly better than the former (53.70), which implies  the difficulty is extracting targets.

The results of the single-domain setting are shown in Table \ref{result4} (last 3 columns).  The average F1 score of open-domain targeted sentiment analysis on the single domain setting is 34.27, better than that of the multi-domain setting (33.98). Overall, open-domain data do not help improve the performance of the model. The worse results for the multi-domain setting (comparing with single-domain setting) are on \textbf{Books} (31.90-34.76) and \textbf{Clothing} (43.14-49.40), which implies that no useful information could be obtained from other domains for these domains. But for \textbf{Restaurant} (20.82-19.08), \textbf{Hotel} (37.12-34.17), \textbf{News} (14.52-12.91) and \textbf{PhraseBank} (56.35-55.27), open-domain data can help boost the  model performance. More effective use of open-domain data requires further research. 




\begin{table}[tp]\small
\centering
\begin{tabular}{c|ccc}
\hline
\hline
            & \textbf{Precision} & \textbf{Recall} & \textbf{F1} \\
\hline
\textbf{1-nest}  &      {47.45}     & \textbf{32.12}       & \textbf{38.31}    \\
\textbf{2-nest}  &      \textbf{50.16 }    &  21.09      &  29.69  \\
\textbf{3-nest} &     29.72      &   20.00     &  23.90  \\
\hline
\hline
\end{tabular}
\caption{Results on different numbers of nests.}
\label{result3}
\end{table}

\subsection{Influence of Document-level Inputs } 
We are interested in understanding the influence of documents for open-domain targeted sentiment analysis, which can be characterized by the average number of tokens or sentences. In particular, we illustrate the relation between the document length and the performance on the multi-domain setting (the illustration on single-domain setting is similar) in Figure \ref{tokens}. The results show that the performance of the model on open-domain targeted sentiment analysis and opinion target extraction has strong correlation to the average number of tokens or sentences, which is one characteristic in the document-level task. With the increase of tokens or sentences, the performance of open-domain targeted sentiment analysis and opinion target extraction decreases significantly. But for target sentiment classification, the performance does not have such an obvious relation (Figure \ref{tokens} (c)(d)) as shown before. This implies that the model can be negatively affected by the document length for open-domain targeted sentiment analysis.
\begin{table}[t]\small

\centering
\begin{tabular}{c|cccc}
\hline
\hline
{\textbf{Domain}} & \textbf{Precision} & \textbf{Recall} & \textbf{F1} \\
\hline
\textbf{Books}        &29.30&12.68&        17.69       \\
\textbf{Clothing}               & 32.47   &   20.72      &    25.29           \\
\textbf{Restaurant}             &23.29&7.98&11.89          \\
\textbf{Hotel}             &27.78&13.00&17.69           \\
\textbf{News}                   & 3.84          &     1.33   &    1.98   \\
\textbf{PhraseBank}             &     33.33  &    25.30     &28.76          \\
\hline
\textbf{Avg} &            25.00&13.50&  17.22       \\
\hline
\hline
\end{tabular}
\caption{Out-of-domain test results (5-1 experiments).}
\label{result6}
\end{table}

\begin{table}[t]\small

\centering
\begin{tabular}{c|ccc}
\hline
\hline
{\textbf{Domain}} & \textbf{Precision} & \textbf{Recall} & \textbf{F1} \\
\hline
\textbf{P->B}        &13.86   &6.14&8.50  \\
\textbf{B->C}               & 16.02 &   15.13      &    15.56          \\
\textbf{C->R}         &7.19&1.97&3.09        \\
\textbf{R->H}         &28.99&13.58&18.49          \\
\textbf{H->N}         &1.87&1.62&1.73           \\
\textbf{N->P}        & 38.07&31.02&34.18    \\
\hline
\textbf{Avg} &   17.67&11.08     &13.59      \\
\hline
\hline
\end{tabular}
\caption{Out-of-domain  test results (1-1 experiments) (P for PhraseBank, B for Books, C for Clothing, R for Restaurant, H for Hotel and N for News).}
\label{result7}
\end{table}


\subsection{Influence of Complex Target Structure }
According to the results shown in Table \ref{result3}, the F1 scores of the 1-nest, 2-nest, and 3-nest settings are 38.31  29.59 and 23.9, showing that the number of target nests negatively affects the performance. The F1 score in the 3-nest target setting is 14.41 lower than that in 1-nest targets experiment.  It implies another reason why the performance of \textbf{Restaurant} (with a large number of 2-nest targets (2566)) is weak.  Nested targets are challenging to identify which requires more inference for the relations between target components for open-domain targeted sentiment analysis.

\begin{figure}[t]
      \centering
      \includegraphics[width=0.95\hsize]{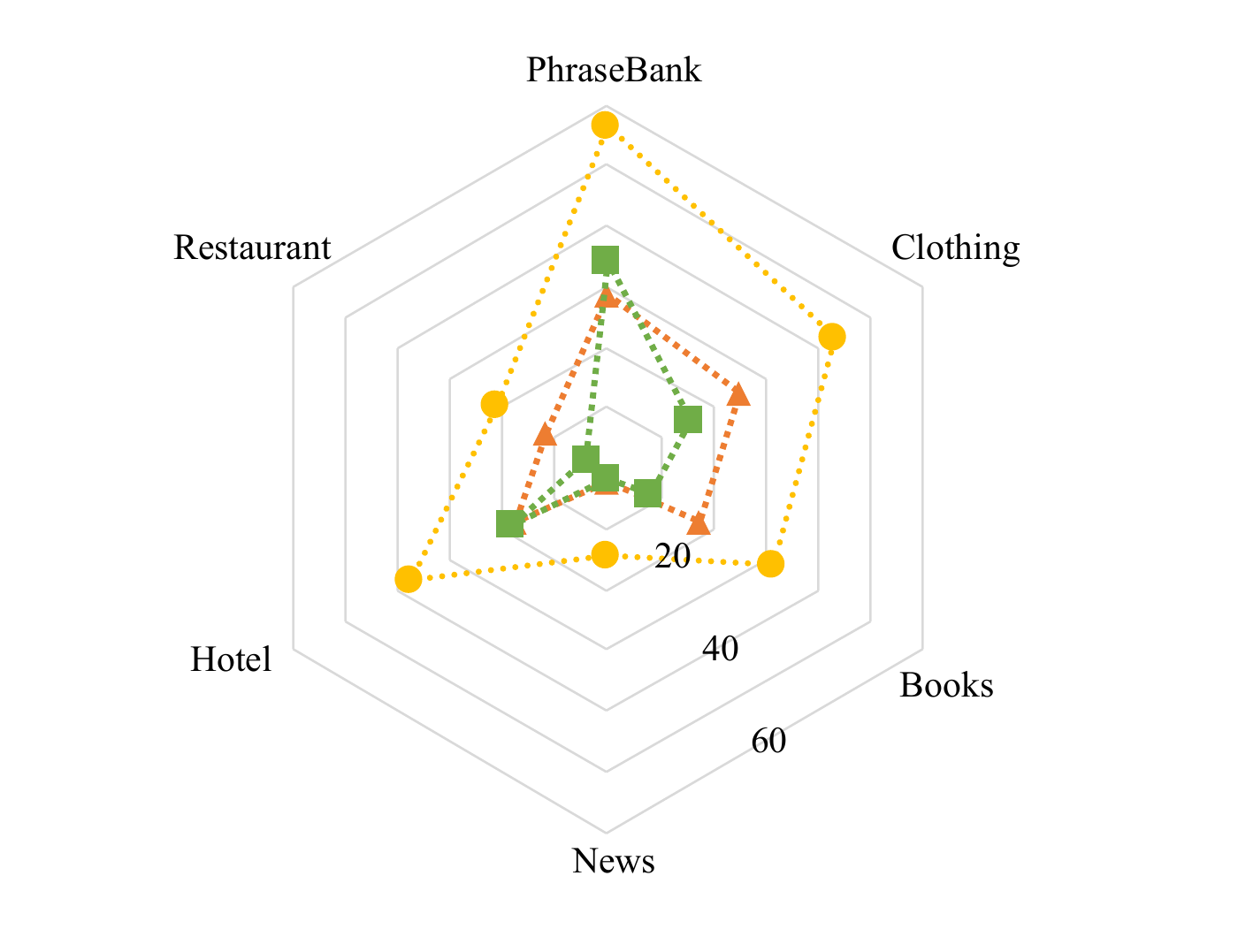}
      \caption{Comparisons between out-of-domain tests and the multi-domain setting.  $\bullet$ symbol for F1 score of the multi-domain setting, $\blacktriangle$  symbol for F1 score of 5-1 out-of-domain test and $\blacksquare$ symbol for F1 score of 1-1 out-of-domain test.}
      \label{Trans}
   \end{figure}
   
   \begin{table}[t]\small

\centering
\begin{tabular}{c|ccc}
\hline
\hline
{\textbf{Domain}} & \textbf{Precision} & \textbf{Recall} & \textbf{F1}\\
\hline
\textbf{Books}                   &   35.46    &   {29.94}     &         32.46         \\
\textbf{Clothing}                &    45.00       &   41.44   &  43.14             \\
\textbf{Restaurant}              &      35.60    &   {26.31}    &     {30.25}           \\
\textbf{Hotel}                   & {59.76}    &   {41.82} &    {49.20}       \\
\textbf{News}                    &      17.85     &  {11.29}      &    {13.83}  \\
\textbf{PhraseBank}             &     58.60     &  51.57     &  54.86          \\

\hline
\textbf{Avg}                     &{42.04}&{33.72}& {37.29}   \\
\hline
\hline
\end{tabular}
\caption{Single-domain setting results of pipeline model.}
\label{result2}
\end{table}

\subsection{Influence of Domain}
The results of 5-1 out-of-domain test are shown in Table \ref{result6}. In particular, the average F1 scores is 17.22, which is 16.75  lower than that on the multi-domain setting. The performance decay implies the generalization performance of the model on our dataset is weak, due to the fact that much difference exists between the domains. The results of 1-1 out-of-domain test are shown in Table \ref{result7}. The average F1 scores of 1-1 out-of-domain test is 14.59, which is 20.38 lower than that on the multi-domain setting, also lower than that on the 5-1 setting. It suggests open-domain data can help to boost the performance of generalization.
The visualization of results in the 5-1 test, 1-1 test and the multi-domain setting is shown in Figure \ref{Trans}.  The performance on the \textbf{News} domain (1.98 and 1.73 in 5-1 and 1-1 tests) is especially low, that the model can hardly learn useful knowledge from other domains for news domain. Note that the results on 1-1 out-of-domain test are better than that on 5-1 test in \textbf{Hotel} (18.39-17.69) and \textbf{PhraseBank} (34.18-28.76), which implies that  more open-domain data does not always lead to  better-trained models.

 \begin{table*}[thp]\small
  \centering
\begin{tabular}{p{0.5\textwidth}p{0.21\textwidth}p{0.21\textwidth}}
\hline
\hline
{\textbf{Context}} & \textbf{Gold Labels} & \textbf{Output} \\
\hline

I've ordered similar character shoes from other manufacturers and, as long as I size up. They fit almost perfectly... perhaps a tad big but a 7 would probably have been too snug. My dissatisfaction is with the strap. Even at the tightest supplied hole, it 's way too loose. & shoes --- strap \# {\color{teal}Negative} $e_{se}$ shoes \# {\color{orange}Mixed} $e_{se}$    & strap \# {\color{teal}Negative} $e_{se}$   \\
\hline
Valerie 's place is spotless with a wonderful kitchen. The only thing that might be difficult for some is the need to climb 2 flights of stairs to access the bedroom. I would stay here again without hesitation. 
&  Valerie 's place --- stairs \# {\color{teal}Negative} $e_{se}$ Valerie 's place \# {\color{orange}Mixed} $e_{se}$  Valerie 's place --- kitchen \# {\color{red}Positive} $e_{se}$ &  Valerie 's place \# {\color{orange}Mixed} $e_{se}$ Valerie 's place --- kitchen \# {\color{red}Positive} $e_{se}$ \\ 
\hline
\hline
\end{tabular}
\caption{Case Study. The symbols '----', '\#' and $e_{se}$ represent the split, ending tokens of target  and the ending token of sentiment, respectively. The beginning token of targets is neglected here for simplicity.}
\label{case}
\end{table*}

\begin{figure}[t]
      \centering
      \includegraphics[width=0.95\hsize]{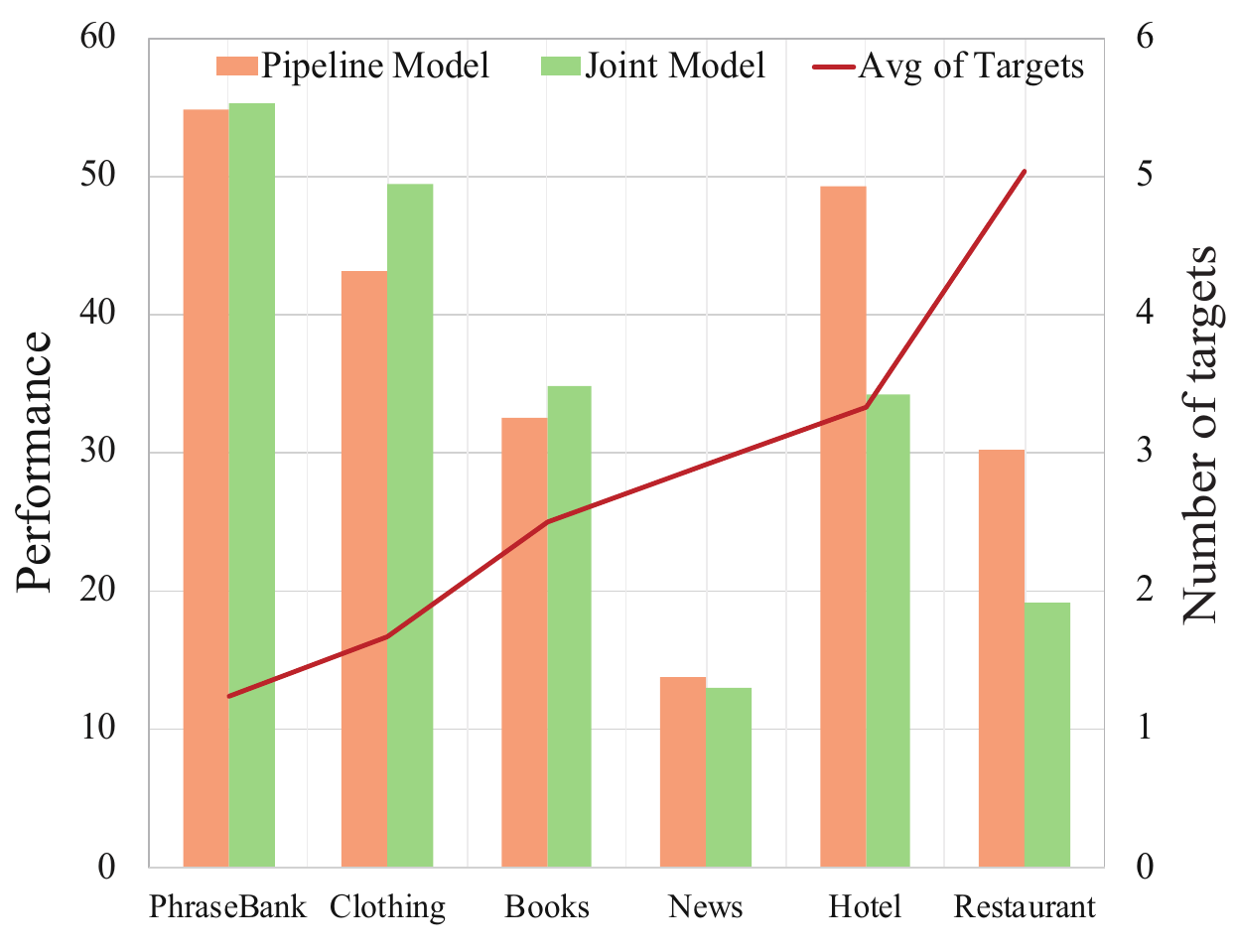}
      \caption{Comparisons between performance of pipeline model and joint model on the single-domain setting.}
      \label{jointpipe}
   \end{figure}

\subsection{Pipeline vs Joint Models}
 Different from the observation of \citet{Mitchell2013}, \citet{Zhang2015} and \citet{Hu2020}, the average F1 score of the pipeline model (37.29) is better than the joint model (34.27). Better results of pipeline  models (comparing with joint models) lie in the domains \textbf{Restaurant} (30.25-19.08), \textbf{Hotel} (39.20-37.12) and \textbf{News} (13.83-12.91). We notice the performance of the joint model is strongly related to the average number of targets in the dataset (Figure \ref{jointpipe}). With the increase of the average number of targets, the performance of the joint model becomes worse than the pipeline model. In the domains that the average number of targets is small (\textbf{Books} (2.50), \textbf{Clothing} (1.67), \textbf{PhraseBank} (1.23)), joint models performer better than pipeline models. Conversely,  in the domains that the average number of targets is large (\textbf{Restaurant} (5.03), \textbf{Hotel} (3.33), \textbf{News} (2.91)), pipeline models have better performance. The phenomenon may be due to the complexity of the generation content, i.e. with the increase of the length of outputs, it becomes harder to generate correct texts for open-domain targeted sentiment analysis, but  opinion target extraction is relatively easier. 

\subsection{Case Study}
Table \ref{case} shows two qualitative cases from the single-domain setting. As observed in the first case, the model outputs a partially correct answer (\textit{strap\#Negative}), but the information of the relation between \textit{shoe} and \textit{strap} is not extracted. Although the words `\textit{They fit almost perfectly}' and `\textit{it's way too loose}' express sentiments for the target \textit{shoe}, it is not extracted, which means that the model fails to infer the anaphora of `\textit{They}' and `\textit{it}'. In the second case, \textit{Valerie 's place --- stairs \#Negative} fails to be extracted when the model faces a relative large number of targets.

\section{Related Work}

\textbf{Open-domain targeted sentiment analysis} can be divided into two sub-tasks, namely, the opinion target extraction and target sentiment classification. Traditionally, the sub-tasks are solved separately \cite{2002Conditional,Lei2017,Zhang2016,ren2016improving,Wang2017,Peng2017,Fan2018,Song2019}, which can be pipelined together to solve the open-domain targeted sentiment analysis.
The joint task of open-domain targeted sentiment analysis is modeled as an end-to-end span extraction problem \cite{Zhou2019,Hu2020} or span tagging problem: tagging as  \textit{\{B, I, E, S\} }- \textit{\{POS, NEG, NEU\}} and \textit{O} \cite{Mitchell2013,Zhang2015,Ma2018,Li2018b,Song2019,Pingili2020}. Recent work compares pipeline model and joint model \cite{Mitchell2013,Zhang2015,Hu2020}, finding that the pipeline model can achieve better performance.



 Previous studies mainly conduct experiments on three datasets: (1) \textbf{LAPTOP},  product reviews from the laptop domain in SemEval 2014 challenge \cite{pontiki2014}; (2) \textbf{TWITTER}, comprised by the tweets collected by Mitchell \cite{Mitchell2013};
(3) \textbf{REST}, a union of restaurant reviews in SemEval 2014, 2015, or 2016 \cite{pontiki2014,pontiki2015,pontiki2016}.  Some work also tries to propose  datasets in news domain \cite{Hamborg2021a,Hamborg2021} which are mainly on the political spectrum. \citet{Orbach2020} constructs a new evaluation dataset in extensive domains finding that there is ample room for improvement on this challenging new dataset.

\textbf{Aspect-based sentiment analysis} is a similar work, which aims to extract the aspect term and then identify its sentiment orientation, like \cite{Li2019,Chen2020,wang2019,Chen2020b,Liu2020}. The task needs to find the aspects related to the elements in a given aspect category set. But for open-domain targeted sentiment analysis, no pre-defined aspect categories are given.  For example in \textbf{LAPTOP} dataset, '\textit{But the \textit{performance} of Mac Mini is a huge disappointment.'} For the target `\textit{{Mac Mini}'} is not in the focused aspect categories, thus it is not labeled and only `\textit{{performance}}' is labeled. Some work tries to extract the target, aspect and sentiment at the same time  \cite{Yang2019,Saeidi2016}, while it limits the extensibility. Meanwhile, document-level aspect-based sentiment analysis task is also studied in \cite{Chen2020,wang2019} to alleviate the information deficiency problem for the implicit targets (pronouns).

\section{Conclusion}
In this study, we propose a challenging dataset for open-domain targeted sentiment analysis. As a solution to the challenging joint and pipeline tasks, we considered a single unified baselines using a seq2seq pre-trained language model, which is close to real-world practical settings. Benchmark performance demonstrated that the task is very difficult even given the current pre-trained technologies, and challenges exist in the effective use of open-domain data, long documents, the complexity of target structure and domain variances.

\section{Ethical Statement}
We honor the Code of Ethics. No private data or non-public information was used in this work. All annotators have received labor fees corresponding to the amount of their annotated instances. 

\bibliography{acl_latex.bib}
\bibliographystyle{acl_natbib}

\clearpage
\appendix

\section{Appendix: Rules for Annotation}
\label{sec:appendix}
\subsection{Target Candidates and sentiment Annotation
}

\vspace{2mm}
\textbf{General Instructions.}

In this task you will review a set of documents. Your goal is to identify the nested items in the documents that have a sentiment expressed to them.

\vspace{2mm}

\noindent \textbf{Steps}

1.	Read the documents thoroughly and carefully.

2.	Identify the items that have a sentiment expressed to them. 

3.	Mark each item by the form of nested target structure connected by ‘--’ and for each nested target choose the expressed sentiment:

(a). {\color{red}Positive}: the expressed sentiment is positive.

(b). {\color{teal}Negative}: the expressed sentiment is negative.

(c). {\color{orange}Mixed}: the expressed sentiment is both positive and negative.

4.	If there is no item with a sentiment expressed towards them, proceed to the next document.

\vspace{2mm}

\noindent \textbf{Rules and Tips}
\begin{enumerate}
    \item The nest target structures are labeled as they appear in the document, even though they have overlapping parts (see example 2).
    \item If the target of pronoun (it, this, that, etc.) could not be inferred from the whole text, the pronoun will be a target, but it will not be considered as a part of nested target structure (see example 2). 
    \item The sentiment should be expressed towards the marked items, it cannot come from with the marked item (see example 3).
    \item Unfactual content will not be marked in conditional or subjunctive sentences (see example 5).
    \item Verbs will not serve as targets even though there exist sentiment words towards them (see example 6).
    \item “the” cannot be a part of a marked item. (see example 7).
    \end{enumerate} 

\subsection{Examples}

\noindent\textbf{1.	Basics}

\vspace{2mm}

\textbf{Example 1.1:} \textit{The food is good.}

\textbf{Answer:} food \# {\color{red}Positive}

\textbf{Explanation:} The word good expresses a positive sentiment towards food.

\vspace{2mm}

\textbf{Example 1.2:} \textit{The food is awful.}

\textbf{Answer:} food \# {\color{teal}Negative}

\textbf{Explanation:} The word awful expresses a negative sentiment towards food.

\vspace{2mm}

\textbf{Example 1.3:} \textit{The food is tasty but expensive.}

\textbf{Answer:} food \# {\color{orange}Mixed}

\textbf{Explanation:} The word good expresses a positive sentiment towards food while the word awful expresses a negative sentiment towards food. So the correct sentiment to food is mixed.

\vspace{2mm}

\textbf{Example 1.4:} \textit{The restaurant is near downtown.}

\textbf{Answer:} Nothing should be selected, for there is no sentiment expressed.

\vspace{2mm}

\noindent\textbf{2.	Nested target structure}

\vspace{2mm}

\textbf{Example 2.1:}\textit{ Good charger and is perfect because it also has a USB connection. Also love that it is original material it works like that too giving a quick charge when I need it.}

\textbf{Answer:} 
charger \# {\color{red}Positive}

charger - USB connection \# {\color{red}Positive}

charger - material \# {\color{red}Positive}

charger - charge \# {\color{red}Positive}

\textbf{Explanation:} The word good expresses a positive sentiment towards charger, and the word perfect expresses a positive sentiment to the USB connection of charger. Meanwhile, the next sentence has positive sentiments towards material and charge separately, and they can be inferred to be a part of the charge.

\vspace{2mm}

\textbf{Example 2.2: } \textit{ It charges my phone quickly and the cord is super long.}
	
\textbf{Answer:}
	It \# {\color{red}Positive}
	
	cord \# {\color{red}Positive}
	
\textbf{Explanation:} The word quickly expresses a positive sentiment to the target it while it cannot be inferred what it represents, then it is marked. For cord, although we can know cord is a part of it, but it will not be considered to be marked in the nested target structure. 

\vspace{2mm}

\textbf{Example 2.3:} \textit{The food was served good for a meal in the Italian restaurant, but the atmosphere was awful. }

\textbf{Answer:}
Italian restaurant - food \# {\color{red}Positive}
	
	Italian restaurant - atmosphere  \#  {\color{teal}Negative}
	
Italian restaurant \#  {\color{orange}Mixed}

\textbf{Explanation:} The word good expresses a positive sentiment to the target food and food is a part of the Italian restaurant, meanwhile for the item food has marked, the duplicate item meal will not be marked. Then word awful expressed a negative opinion towards the atmosphere of the Italian restaurant. Further, from the two nested target items, the sentiment of Italian restaurant can be inferred to be mixed.

\vspace{2mm}

\noindent\textbf{3.	Sentiment location}

\vspace{2mm}

\textbf{Example 3.1:} \textit{I love this great car.}

\textbf{Answer:} car \# {\color{red}Positive}

\textbf{Explanation:} Both words love and great expresses positive sentiment towards car, so car is marked, but not great car is marked.

\vspace{2mm}

\noindent\textbf{4.	Long-document examples}

\vspace{2mm}

\textbf{Example 4.1:} \textit{Could not power my S2 phone. The LG charger I was using had no problem but I needed a second charger.  I thought buying an Official Samsung charger would be the best route to go. With nothing running on my phone except Waze and Audible (my usual combo when driving) the battery icon showed charging on AC, BUT was losing power at the rate of 5\% per hour. On a long trip I was forced to turn the phone completely off for a few hours to get it to charge. In fairness it could have been a defective unit but I won't be wasting time trying another of this model. The company has been very accommodating in the return. The return has been smooth and I WOULD buy from them again.}

\textbf{Answer:}
LG charger \# {\color{red}Positive}

Official Samsung charger \# {\color{teal}Negative}

company \# {\color{red}Positive}

company--return \# {\color{red}Positive}






\vspace{2mm}

\textbf{Example 4.3}: \textit{My wife liked my Nokia 3650 so much that she switched chips with me and is carrying it. My favorite features:1.  Speaker Phone. Nice when driving or multitasking. Good audible range. I slip it in my shirt pocket and speak into the air. Works great!2.  Display is very good for its size. The camera takes 640 x 480 color images. I bought a 32 meg card to increase storage. I recently used the phone as my principle camera on vacation to the Smokies. Worked great.3. Contacts is a nice feature that can pull your chip's phone numbers and store them. Just add email addresses and you can send the camera pics to any email via the multimedia option. Disadvantages:  The blue lighted round keyboard. In low light it is hard to see. This can be a problem when text-messaging or adding contact details. I'm buying a 2nd phone which will be another Nokia 3650. (...) :)}

\textbf{Answer:}
Nokia 3650 \# {\color{orange}Mixed}

Nokia 3650--Speaker \# {\color{red}Positive}

Nokia 3650--Speaker--audible range \# {\color{red}Positive}

Nokia 3650--display--size \# {\color{red}Positive}

Nokia 3650--camera \# {\color{red}Positive}

Nokia 3650--contact \# {\color{red}Positive}

Nokia 3650--multimedia option \# {\color{red}Positive}

Nokia 3650--keyboard \# {\color{teal}Negative}

\vspace{2mm}

\noindent\textbf{5.	Unfactual content will not be marked in conditional or subjunctive sentences}

\vspace{2mm}

\textbf{Example 5.1:} \textit{For example, if the Asia Pacific market does not grow as anticipated, our results could suffer.}

\textbf{Answer:} Nothing should be selected, for the sentence is a conditional sentence.

\vspace{2mm}

\noindent\textbf{6.	Verbs not for targets}

\vspace{2mm}

\textbf{Example 6.1:} \textit{Works well. }

\textbf{Answer:} Nothing should be selected, for verbs will not be targets. It is normal to be marked in the ABSA work, for they can be aspects of the items.

\vspace{2mm}

\noindent\textbf{7.	“the” cannot be a part of a marked item}

\vspace{2mm}

\textbf{Example 7.1:} \textit{The food is awful.}

\textbf{Answer:} food \# {\color{teal}Negative}

\textbf{Error:} The food \# {\color{teal}Negative}

\vspace{2mm}

\noindent\textbf{8.	Idioms}

\vspace{2mm}

\textbf{Example 8.1: } \textit{The laptop’s performance was in the middle of the pack, but so is its price.}

\textbf{Answer:} None 

\textbf{Explanation:} A sentiment may be conveyed with an idiom – be sure you understand the meaning of an input sentence before answering. When unsure, look up potential idioms online. in the middle of the pack does not convey a positive nor a negative sentiment, and certainly not both (so the answer is not "mixed" as well).

\section{Appendix: Data Source}
 Our proposed dataset contains six domains, including books reviews, clothing reviews, restaurant reviews, hotel reviews, financial news and social media data. 

\subsection{Dataset Sources}
Raw document data are from several datasets or collected by ourselves and they are used for annotation inputs. The details are as follows:

\begin{enumerate}
\item \noindent\textbf{Books and Clothing.} The reviews of books and clothing are from \footnote{\url{https://nijianmo.github.io/amazon/index.html}}. 
The annotated data contains 986 book reviews and 928 clothing reviews which are randomly selected from the downloaded dataset.
We used the data of books domain and clothing domain  of 5-core version in this data source.
\vspace{-3mm}

\item \noindent\textbf{Restaurant.} Restaurant reviews are in Boston, collected by Yelp (April 17, 2021).\footnote{\url{https://www.yelp.com/dataset/download}}  The annotated data contains 940 reviews which are randomly selected from the downloaded dataset (only restaurant reviews remain).

\vspace{-3mm}

\item \noindent\textbf{Hotels.} Hotel reviews are in Boston, collected by AirBnb (February 19, 2021).\footnote{\url{http://insideairbnb.com/get-the-data.html} } The annotated data contains 1029 reviews which are randomly selected from the downloaded dataset.

 \vspace{-3mm}

\item \noindent\textbf{Social Media.}  A random sample of 1194 sentences was chosen to represent the overall social media database\footnote{\url{https://huggingface.co/datasets/financial_phrasebank}}.
Annotators were asked to consider the sentiment of sentences from the view point of an investor only.

\vspace{-3mm}
    
\item \noindent\textbf{Business News.} Our business news dataset was collected from Reuters\footnote{\url{https://www.reuters.com/news/}} and Bloomberg\footnote{\url{https://github.com/philipperemy/financial-news-dataset}} containing 936 news. In particular, Reuters News was collected from March 2021 to April 2021, resulting in 498 instances. While Bloomberg News was collected over the period from October 2006 to November 2013, resulting in 438 samples. 
\end{enumerate} 
\newpage
\subsection{Data Statistics}
 The numbers of targets in the different number of target nests are shown in Table 8. It shows that most of the targets are 1-nest and 2-nest, some are 3-nest and few are 4-nest.

\begin{table}[h]\small
\begin{tabular}{cccccc}   
\hline
\hline
Domain &1-nest   & 2-nest & 3-nest& 4-nest\\
\hline
Books &1465&988&17&0\\
Clothing &1166&385&4&0\\
Restaurant &1943&2566&221&9\\
Hotel &1408&1795&231&2\\
News& 2053 &618&53&1 \\
PhraseBank &918&541&49 &0\\
\hline
\hline
\end{tabular}
\caption{The statistics of target nests in proposed dataset.}
\end{table}

\end{document}